\documentclass[pmlr]{jmlr}

\usepackage{amsmath,amssymb,graphicx,url}
\usepackage{longtable}
\usepackage{booktabs}
\usepackage{siunitx}
\usepackage[edges]{forest}
\usepackage{float}
\usepackage{nicefrac}

\usepackage{pgfplots}
\usetikzlibrary{spy}
\usetikzlibrary{patterns}
\usepgfplotslibrary{groupplots, statistics}
\pgfplotsset{compat=1.18}

\definecolor{cvblue}{RGB}{31,119,180}
\definecolor{jabred}{RGB}{255,127,14}
\definecolor{splitpurple}{RGB}{148,103,189}
\definecolor{evalteal}{RGB}{105,126,140}
\definecolor{evalmustard}{RGB}{201,82,107}
\definecolor{evalsky}{RGB}{184,115,51}
\definecolor{evalmagenta}{RGB}{23,138,148}
\definecolor{wforest}{RGB}{44,160,44}
\definecolor{wrose}{RGB}{214,39,40}
\definecolor{wslate}{RGB}{127,127,127}
\definecolor{wgold}{RGB}{128,54,128}
\definecolor{wforestPlum}{RGB}{128,54,128}
\definecolor{wforestCoral}{RGB}{205,96,86}
\definecolor{wforestCream}{RGB}{245,242,232}
\definecolor{wforestCharcoal}{RGB}{40,45,42}

\usepackage[cachedir=minted-cache,highlightmode=immediate]{minted}

\usepackage{tcolorbox}
\tcbuselibrary{listings, minted, skins, breakable}

\usepackage{csquotes}

\newtcblisting{pythoncode}{
  listing only,       
  colback=black!3,
  colframe=black,
  listing engine=minted,
  minted language=python,
  minted options={fontsize=\small, linenos},
  left=5pt, right=5pt, top=5pt, bottom=5pt
}

\newtcolorbox{warningbox}[1][]{
    colback=red!5!white,     
    colframe=red!75!black,   
    fonttitle=\bfseries,     
    title=Warning,           
    boxrule=1pt,             
    arc=4pt,                 
    #1
}

\theorembodyfont{\upshape}
\theoremheaderfont{\scshape}
\theorempostheader{:}
\theoremsep{\newline}

\jmlrvolume{329}
\jmlryear{2026}
\jmlrworkshop{Conformal and Probabilistic Prediction with Applications}
\jmlrproceedings{PMLR}{Proceedings of Machine Learning Research}

\title[Conformal Anomaly Detection in Python]{Conformal Anomaly Detection in Python:\\
Moving Beyond Heuristic Thresholds with \texttt{nonconform}}

\author{\Name{Oliver Hennh\"ofer} \Email{oliver.hennhoefer@h-ka.de}\\
\Name{Maximilian Kirsch} \Email{maximilian.kirsch@h-ka.de}\\
\Name{Christine Preisach} \Email{christine.preisach@h-ka.de}\\
\addr Intelligent Systems Research Group, \\ Karlsruhe University of Applied Sciences, Germany}

\editor{Ernst Ahlberg, Ulf Johansson, Henrik Boström, Alberto Carlevaro, Johan Hallberg Szabadváry and Lars Carlsson}

\begin{document}

\maketitle

\begin{abstract}
Most anomaly detection systems output scores rather than calibrated decisions, leaving practitioners to choose thresholds heuristically and without clear statistical interpretation. Conformal anomaly detection addresses this limitation by converting anomaly scores into calibrated \textit{p}-values that are valid under the statistical assumption of data exchangeability, with a growing literature extending this idea beyond that setting. We present \texttt{nonconform}, a Python package for applying conformal anomaly detection within existing machine-learning workflows, and use it as the basis for an implementation-grounded introduction to the field. The package integrates with \texttt{scikit-learn}, \texttt{PyOD}, and custom anomaly detectors, and provides a unified interface for calibration, \textit{p}-value generation, and false discovery rate control. It supports several conformalization strategies, ranging from simple split-conformal calibration to more data-efficient and shift-aware extensions. Through a progression from foundational concepts to advanced conformalization strategies, complemented by code examples, the paper connects the statistical ideas behind conformal anomaly detection to their practical use in \texttt{nonconform}. Empirical results demonstrate that the implemented methods enable statistically principled anomaly detection. Together, the package and exposition aim to make core conformal anomaly detection workflows more accessible and reproducible in experimental and production-oriented settings.
\end{abstract}


\section{Introduction}
\label{sec:intro}
Anomaly detection is fundamentally an exercise in drawing a line between the \enquote{expected}, which \textit{conforms} to an anticipated state of normality, and the \enquote{exceptional}. In practice, this task is often delegated to scoring functions such as the isolation path length in an \textit{Isolation Forest} \citep{Liu2008} or the distance to a hyperplane in a \textit{One-Class Support Vector Machine} \citep{Schoelkopf2001}. However, while these models are highly effective at ranking observations by their degree of anomaly, the resulting scores are typically heuristic and lack an intrinsic statistical interpretation. A practitioner who wants to move from an anomaly score to a decision must still choose a threshold --- often using intuition, visual inspection, or more elaborate procedures that remain only loosely connected to the actual reliability of the resulting decisions. In other words, such ad hoc approaches offer no formal control of false alarms, leaving practitioners unable to answer a basic operational question: \enquote{\textit{If I use this threshold, how often will I flag normal observations by mistake?}}

Without a calibrated link between threshold choice and error rates, deploying anomaly detection systems becomes difficult in settings where decisions must be operationally reliable and statistically interpretable. \textbf{Conformal Anomaly Detection (CAD)} \citep{Laxhammar2015,Bates2023} addresses this limitation by calibrating raw anomaly scores into valid \textit{p}-values that quantify how unusual a new observation appears relative to a reference sample of normal data. In this way, CAD turns anomaly detection from a purely ranking-based task into a statistically interpretable testing problem, enabling decisions to be based on established statistical procedures with explicit control over false alarms.

\subsection{Motivation: Statistical Error Control}
\label{subsec:control}

In many real-world anomaly detection tasks, the cost of a false positive is a direct operational risk. Beyond the immediate cost of investigation, a high rate of false positives can trigger \textit{alert fatigue} and erode trust in the underlying system. For an anomaly detection system to be operationally reliable, a practitioner should be able to specify a tolerated nominal error rate $\alpha \in (0,1)$ (e.g. $\alpha=0.05$, or 5\%) and expect the resulting decision rule to respect this level in the long run. The appropriate choice of $\alpha$ depends on the application, since tolerance for false alarms is tied to the consequences of acting on them.

Standard anomaly detection methods do not usually produce outputs whose thresholds admit such a direct interpretation. As a result, the relationship between a chosen score threshold and the resulting false alarm rate is typically unclear until the system is deployed.

Under the standard conformal framework, this interpretability takes a concrete form. If a test observation is flagged whenever its conformal \textit{p}-value is at most $\alpha$, then under the null hypothesis $\mathcal{H}_0$ that the observation is an inlier, and under the assumptions required for conformal validity,
\[
\mathbb{P}(\text{False Positive}) \le \alpha.
\]
This guarantee is largely independent of the particular anomaly scoring model used and therefore provides a robust statistical foundation for decision-making.

\subsection{The Challenge of Scale: Multiple Testing}

Controlling false alarms for a single observation is an important first step, but practical anomaly detection systems rarely evaluate observations in isolation. In monitoring, screening, or streaming settings, models may assess observations at scale, either in batches or sequentially. Once each observation is associated with its own statistical test, anomaly detection becomes a multiple-testing problem \citep{Shaffer1995}. A decision rule calibrated at $\alpha = 0.05$ per instance implies that, on average, at most one in twenty normal observations will be flagged purely by chance. In high-throughput settings, these individual risks accumulate, so that even when each decision is controlled in isolation, the overall number of false alarms can become substantial.

A natural reaction is to simply lower the tolerated error rate, making each individual test more conservative. While this reduces false alarms, it also suppresses genuine anomalies. More fundamentally, controlling errors \textit{per observation} does not tell a practitioner much about the quality of the \textit{set} of observations that is ultimately flagged. Operationally, the more relevant question is often: \enquote{\textit{Among all alerts, what fraction turns out to be false alarms?}}

Informally, this is the quantity captured by the \textbf{False Discovery Rate (FDR)} \citep{Benjamini1995}, which can be thought of as

\[
\text{FDR} \approx \frac{\text{False Alarms}}{\text{Total Alarms}}
\quad\text{or, operationally,}\quad
\frac{\text{Wasted Investigation Effort}}{\text{Total Investigation Effort}}.
\]

Rather than constraining how often any one normal observation is mistakenly flagged, FDR controls the expected proportion of false positives among all flagged observations. In practice, this makes FDR control not merely a statistical safeguard but a resource-management principle: it allows some false alarms while ensuring that the selected set of anomalies remains informative overall.

Standard procedures for FDR control, most notably the \textbf{Benjamini--Hochberg Procedure (BH)} \citep{Benjamini1995}, operate on a collection of \textit{p}-values and determine which observations to flag while controlling the FDR at a chosen target level. Their practical effect is to filter a potentially large set of candidate anomalies down to a smaller and more reliable discovery set. Observations that would pass a naive per-instance threshold may no longer survive this multiple testing correction, while sufficiently extreme cases remain selected. This selective behaviour is what makes FDR control well-suited for explorative tasks like anomaly detection at scale. While BH is defined for batch settings, analogous procedures also exist for sequential testing and other testing frameworks, e.g. \citet{Vovk2003}, \citet{Vovk2021} or \citet{Javanmard2018}.

\subsection{From Scores to Decisions}

The multiple-testing perspective described above relies on the availability of valid \textit{p}-values. Standard anomaly detectors do not produce them directly. CAD bridges this gap by comparing the anomaly score of a new observation with scores computed on reference data known to be normal. If the new score is more extreme than most reference scores, this provides evidence that the observation does not conform to the expected pattern. In that sense, CAD turns the raw output of a scoring model into a valid statistical \textit{p}-value.

In the standard setting, the validity of this $p$-value construction rests on the statistical assumption of \textbf{Exchangeability} \citep{Vovk2005}. Informally, this means that the calibration observations and the new test observation can be regarded as being generated symmetrically under the null, so that none occupies a privileged position. A key consequence is that the statistical guarantee attaches to the calibration procedure rather than to the anomaly detector itself. CAD is therefore entirely agnostic to the underlying model.

That said, exchangeability is not always a realistic assumption in practice. Temporal data naturally exhibit serial dependence, and spatial data often show systematic spatial dependence. In both cases, the order or location of observations carries information, which violates the symmetry required by the standard conformal framework. This should not be understood as a limitation of conformal methods in general, but rather of their classical formulation. A growing body of research has developed extensions that relax the exchangeability assumption and recover approximate or exact validity guarantees in settings where the standard approach no longer applies directly. Where such extensions are applicable to anomaly detection, they are implemented in \texttt{nonconform} as the library evolves. The present work focuses on more classical conformal methods, with the aim of developing the foundational concepts, assumptions, and validity arguments that underpin the field. 

\subsection{Scope and Outline}

As outlined, CAD serves two closely related purposes. It gives anomaly scores a statistical interpretation as calibrated \textit{p}-values and consequently enables principled error control with statistical guarantees in downstream tasks. In this context, the paper develops different perspectives on workflows and outcomes while introducing \texttt{nonconform}.

Section~\ref{sec:cad} presents the theoretical and methodological foundations of CAD, beginning with the standard framework and then extending to more advanced  strategies, including data-efficient resampling-based variants and methods designed for settings beyond the classical exchangeability assumption. Section~\ref{sec:martingale} broadens the perspective by introducing conformal martingales as an advanced sequential use of conformal \textit{p}-values, with emphasis on their interpretation as tools for change-point detection. Section~\ref{sec:software} provides a technical introduction to \texttt{nonconform}, its design principles, and its integration with the broader Python ecosystem. Section~\ref{sec:conclusion} concludes with an outlook for the future direction of this project.

\paragraph{Demonstration.} Each section is accompanied by empirical results to build intuition on the observed behaviour and potential failure modes. As practical demonstrations rather than a comprehensive empirical evaluation, all experiments use scikit-learn’s Isolation Forest \citep{Liu2008} on the UCI Statlog (Shuttle) dataset \citep{Shuttle}.\footnote{All results are reproducible using the code available at \href{https://github.com/OliverHennhoefer/nonconform-paper}{github.com/OliverHennhoefer/nonconform-paper}.}

\section{Conformal Anomaly Detection}
\label{sec:cad}

Section~\ref{sec:intro} motivated CAD as a way to turn heuristic anomaly scores into statistically interpretable decisions. We now formalize that idea in the standard setting. The basic construction is simple: a new observation is assigned an anomaly score; this score is compared with scores computed on reference calibration data, disjoint from the original training data, that is known to be normal. The test score's relative rank among the calibration scores is the test point's \textit{p}-value (see Section~\ref{sec:inductive}). Under the exchangeability assumption introduced above, these \textit{p}-values are valid in finite samples, regardless of the particular anomaly detector used to generate the scores.

\subsection{Problem Setup and Notation}
\label{sec:notation}

We begin by fixing notation for the standard setting. Let \(X_1, \dots, X_n \in \mathcal{X}\) denote observations representing normal behaviour, and let \(X_{n+1}, \dots, X_{n+m} \in \mathcal{X}\) denote new observations to be assessed. An anomaly detector is represented by a scoring function
\[
s : \mathcal{X} \to \mathbb{R},
\]
where larger values indicate greater deviation from the reference pattern. The role of CAD is to calibrate these scores so that they admit a statistical interpretation.

To this end, the scores of the test observations are compared with the scores of the reference sample, yielding a \textit{p}-value for each \(X_{n+j}\), \(j=1,\dots,m\). These \textit{p}-values quantify how extreme the test scores are relative to the normal reference observations and can then be used either directly for thresholding or as input to multiple-testing procedures. The case \(m=1\) corresponds to testing a single new observation, as described in Section~\ref{subsec:control}.

\subsection{Inductive Conformal Anomaly Detection}
\label{sec:inductive}

We now make this construction concrete in the standard inductive, or split-conformal, setting: The reference sample \(X_1, \dots, X_n\) is partitioned into a training set \(\mathcal{D}_{\mathrm{train}}\) and a calibration set \(\mathcal{D}_{\mathrm{cal}}\), with
\[
\mathcal{D}_{\mathrm{train}} \cap \mathcal{D}_{\mathrm{cal}} = \varnothing,
\qquad
\mathcal{D}_{\mathrm{train}} \cup \mathcal{D}_{\mathrm{cal}} = \{X_1,\dots,X_n\}.
\]
The scoring function \(s\) is fitted using only \(\mathcal{D}_{\mathrm{train}}\). It is then evaluated on the calibration observations to obtain calibration scores
\[
S_i := s(X_i), \qquad X_i \in \mathcal{D}_{\mathrm{cal}}.
\]

Consider first a single new observation \(X_{n+1}\), with corresponding score
\[
S_{n+1} := s(X_{n+1}).
\]
Under the convention that larger scores indicate greater deviation from normality, the split-conformal \textit{p}-value is defined as
\[
    \hat{p}(X_{n+1})
=
\frac{\sum_{X_i \in \mathcal{D}_{\mathrm{cal}}} \mathbf{1}\{S_i \geq S_{n+1}\} + 1}
{|\mathcal{D}_{\mathrm{cal}}| + 1}.
\]

This quantity is the fraction of calibration scores that are at least as large as the test score, including a finite-sample correction. Small values of \(\hat{p}(X_{n+1})\) indicate that the test score is unusually extreme relative to the calibration set and therefore provides evidence against the null hypothesis that \(X_{n+1}\) follows the same distribution as the normal reference observations.

\paragraph{Validity.}
The importance of this construction is that \(\hat{p}(X_{n+1})\) is a valid \textit{p}-value under the standard conformal assumptions. More precisely, if the calibration observations and the test point are exchangeable, and if the score function \(s\) is fitted using only \(\mathcal{D}_{\mathrm{train}}\), then
\[
\mathbb{P}\!\left(\hat{p}(X_{n+1}) \le t\right) \le t,
\qquad t \in [0,1].
\]
Thus, under the null, the conformal \textit{p}-value is super-uniform, i.e. \textit{at worst} biased towards conservativeness. In particular, for any significance level \(\alpha \in (0,1)\),
\[
\mathbb{P}\!\left(\hat{p}(X_{n+1}) \le \alpha\right) \le \alpha,
\]
so the decision rule that flags \(X_{n+1}\) whenever \(\hat{p}(X_{n+1}) \le \alpha\) controls the type-I error in finite samples. This is the formal version of the operational guarantee discussed in the introduction: once anomaly scores have been conformalized, thresholding at level \(\alpha\) yields a false-alarm probability of at most \(\alpha\) under the null hypothesis.

This guarantee relies on the sample split. Once \(s\) has been fitted on \(\mathcal{D}_{\mathrm{train}}\), the calibration and test scores are computed using the same fixed scoring rule and are therefore comparable under the null hypothesis. The \(+1\) correction in numerator and denominator ensures exact finite-sample validity and prevents \textit{p}-values equal to zero. Since conformal \textit{p}-values lie on a discrete grid, they are generally not exactly uniform, but rather super-uniform.

Accordingly, the standard inductive conformal decision rule flags \(X_{n+1}\) whenever
\[
\hat{p}(X_{n+1}) \le \alpha.
\]

The discriminative performance of \(s\) is not part of the validity argument, but it directly affects statistical power and determines whether and how many discoveries can be made.

\subsection{From Individual Decisions to Multiple Testing}

The single-observation construction extends directly to the batch setting. For test observations \(X_{n+1}, \dots, X_{n+m}\), we obtain conformal \textit{p}-values
\[
\hat{p}_j := \hat{p}(X_{n+j}), \qquad j=1,\dots,m.
\]
Each of these \textit{p}-values is marginally valid under its corresponding null hypothesis. At the same time, because they are all computed using the same random calibration sample, they exhibit dependence stemming from the common calibration set inducing a shared source of randomness across all test points.

When many observations are screened simultaneously, marginal validity of the individual \(\hat p_j\) values is no longer sufficient if one seeks a batch-level error guarantee. In that setting, thresholding the \(\hat p_j\) values separately at an unadjusted level is generally inadequate, and the problem is more naturally formulated as one of multiple testing. One therefore applies a multiple-testing procedure to \((\hat p_1,\dots,\hat p_m)\) to obtain binary decisions \((\delta_1,\dots,\delta_m)\in\{0,1\}^m\), where \(\delta_j=1\) indicates that \(X_{n+j}\) is flagged as anomalous. The mentioned dependence among the conformal \textit{p}-values then becomes important, because it affects which multiple-testing procedures can be justified under the results of \citet{Bates2023}.

A standard choice is the BH procedure, which targets FDR control at a prescribed level \(\alpha \in (0,1)\), even under this dependence. Let
\[
\hat{p}_{(1)} \le \hat{p}_{(2)} \le \dots \le \hat{p}_{(m)}
\]
denote the ordered conformal \textit{p}-values, and define
\[
k^* := \max \left\{ k \in \{1,\dots,m\} :
\hat{p}_{(k)} \le \frac{k}{m}\alpha \right\},
\]
with \(k^*=0\) if the set is empty. If \(k^*>0\), BH flags all
observations satisfying
\[
\hat{p}_j \le \hat{p}_{(k^*)}.
\]
Whether this step enjoys rigorous finite-sample validity depends on the dependence structure of the conformal \textit{p}-values, a point returned to in Section~\ref{sec:wcad}.

This separates the CAD pipeline into two conceptually distinct steps: first, conformal calibration turns raw anomaly scores into valid per-observation \textit{p}-values. Second, a multiple-testing procedure turns those \textit{p}-values into a coherent set of anomaly decisions at the desired error level. In this way, the operational perspective from Section~\ref{sec:intro} carries over to the formal setting: conformalization provides calibrated evidence for each individual observation, while multiple testing determines how that evidence is aggregated when anomaly detection is performed at scale.

When observations arrive sequentially, conformal \textit{p}-values can be tested with online testing variants of the BH procedure. However, in practice these methods are more conservative, since decisions must be made without access to future \textit{p}-values, see \citet{Javanmard2018}.

\begin{tcolorbox}[
  breakable,
  enhanced,
  listing engine=minted,
  minted language=python,
  colback=orange!10,
  colframe=orange!60,
  coltitle=black,
  title=Implementation: Inductive Conformal Anomaly Detection,
  lower separated=true]
\begin{minted}{python}
from oddball import Dataset, load  # Optional dependency
from pyod.models.iforest import IForest

from nonconform import ConformalDetector, Split
from nonconform.metrics import false_discovery_rate, statistical_power

x_train, x_test, y_test = load(Dataset.SHUTTLE, setup=True, seed=42)

detector = ConformalDetector(
    detector=IForest(),
    strategy=Split(1_000),  # Splits calibration set from training set
    seed=42
)

detector.fit(x_train)  # Trains and calibrates

decisions = detector.select(x_test, alpha=0.2)  # Applies BH Procedure

print(f"Empirical FDR: {false_discovery_rate(y_test, decisions)}")
print(f"Statistical Power: {statistical_power(y_test, decisions)}")
\end{minted}

\tcblower
\texttt{Empirical FDR: 0.18}\\
\texttt{Statistical Power: 0.99}
\end{tcolorbox}

 \paragraph{Detached Calibration.}
While the standard conformal pipeline fits the model and calibrates it in a single coordinated sequence, this is not always feasible in production environments where base models are computationally expensive or already deployed. To accommodate this, \texttt{nonconform} supports a \emph{detached} calibration workflow. Practitioners can wrap an already-fitted model and bypass the training phase entirely, applying \texttt{.calibrate(X\_calib)} directly to a held-out inlier set, allowing for retroactive detector conformalization.

\subsection{Resampling-based Extensions}
\label{sec:strategies}

The split-conformal construction offers the clearest route to finite-sample validity, but it requires a calibration set that is disjoint from the data used to fit the anomaly detector. In small samples, this can be costly as fewer observations remain for fitting, and the resulting conformal \textit{p}-values become coarse because they lie on the grid
\[
\left\{\frac{1}{|\mathcal{D}_{\mathrm{cal}}|+1}, \frac{2}{|\mathcal{D}_{\mathrm{cal}}|+1}, \dots, 1\right\}.
\]
In CAD, this discreteness matters directly, since a small calibration set limits the smallest attainable \textit{p}-value and can make downstream testing procedures less powerful. 

This motivates resampling-based extensions \citep{Hennhoefer2024} that reuse the available inlier data more efficiently while preserving the symmetry structure required for conformal validity approximately \citep{Vovk2015}.

Two important families are cross-conformal and bootstrap-based methods. The former partitions the data into \(K\) folds, fits the detector repeatedly on \(K-1\) folds, and uses the held-out fold for scoring, so that every observation receives an out-of-sample score while more data are used for fitting than in split conformal. 

\begin{tcolorbox}[
  enhanced,
  listing engine=minted,
  minted language=python,
  colback=orange!10,
  colframe=orange!60,
  coltitle=black,
  title=Implementation: Cross-Conformal Calibration (CV/CV$+$),
  lower separated=true]
\begin{minted}{python}
detector = ConformalDetector(
    detector=IForest(),
    strategy=CrossValidation(k=10, mode="plus")
)
\end{minted}
\end{tcolorbox}

Leave-one-out (also called \textit{Jackknife}) is the special case with \(K=n\). 

\begin{tcolorbox}[
  enhanced,
  listing engine=minted,
  minted language=python,
  colback=orange!10,
  colframe=orange!60,
  coltitle=black,
  title=Implementation: Leave-One-Out-Conformal Calibration,
  lower separated=true]
\begin{minted}{python}
detector = ConformalDetector(
    detector=IForest(),
    strategy=CrossValidation.jackknife(mode="plus")
)
\end{minted}
\end{tcolorbox}

Bootstrap-based variants instead replace deterministic folds by repeated resampling with replacement and aggregate out-of-bag scores across bootstrap models. In the implemented \enquote{\(+\)} modes, fold- or bootstrap-specific models are retained and their test scores are aggregated, while calibration uses out-of-fold or out-of-bag scores.

\begin{tcolorbox}[
  enhanced,
  listing engine=minted,
  minted language=python,
  colback=orange!10,
  colframe=orange!60,
  coltitle=black,
  title=Implementation: Bootstrap-Conformal Calibration (JaB/JaB+),
  lower separated=true]
\begin{minted}{python}
detector = ConformalDetector(
    detector=IForest(),
    strategy=JackknifeBootstrap(n_bootstraps=100, mode="plus")
)
\end{minted}
\end{tcolorbox}

In general, these methods trade additional computation for better data
efficiency and higher \textit{p}-value resolution than split conformal.
Their validity guarantees are method-specific and are not equivalent to the exact split-conformal guarantee. The non-\enquote{\(+\)} variants additionally rely on stability-type approximations.

\begin{figure}[H]
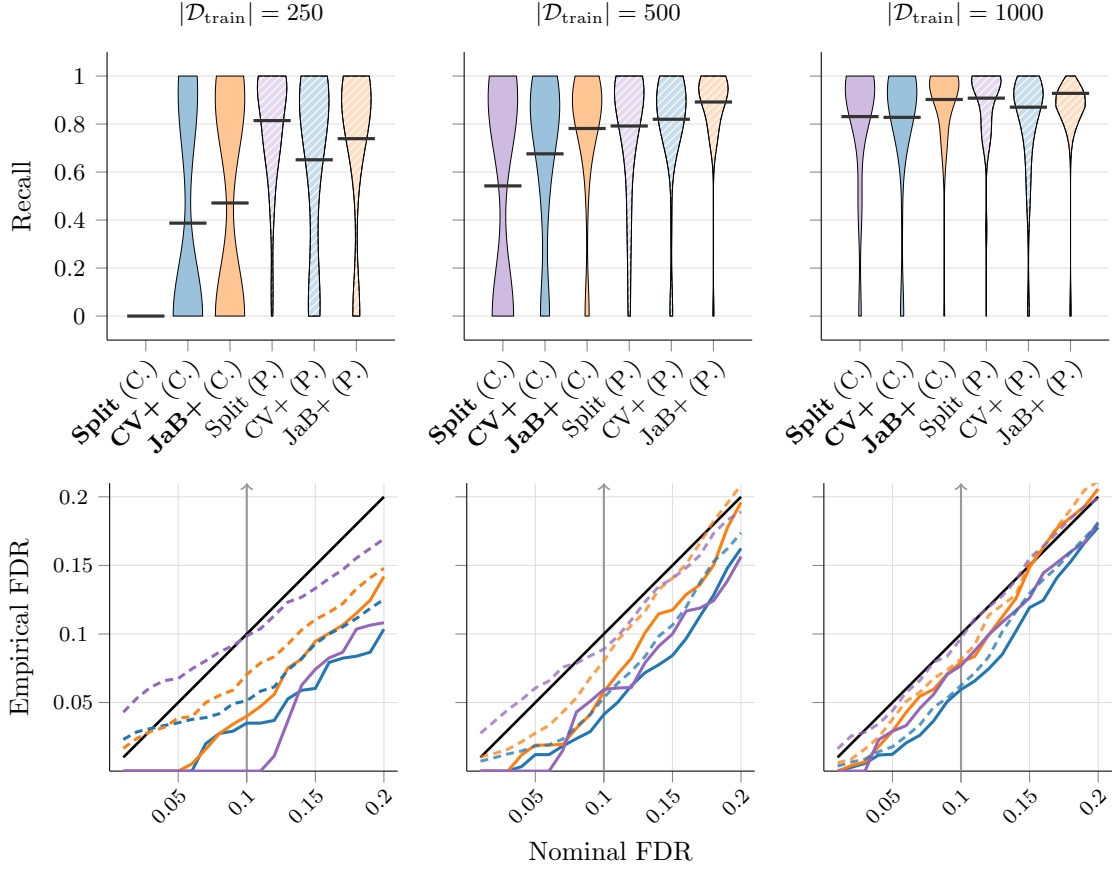

    \centering
    
    \input{figures/1-strategy-plots/cad_violin}

    \vspace{0.5em} 
    
    \input{figures/1-strategy-plots/cad_linegraph}
    
    \caption{\textbf{Recall and FDR depend on the calibration-set size.}
Top: distribution of recall at nominal FDR level $\alpha = 0.1$ for training-set sizes \mbox{$|\mathcal{D}_{\mathrm{train}}| \in \{250, 500, 1000\}$}. Bottom: empirical FDR across corresponding nominal FDR levels. Depicted are the standard conformal and the probabilistic (P.) approach. The \textcolor{splitpurple}{\textbf{Split}} variant is calibrated on $\mathcal{D}_{\mathrm{train}}/2$. Results are averaged over 50 randomized trials. \textcolor{jabred}{\textbf{JaB+}} uses \texttt{n\_bootstraps}=100, and \textcolor{cvblue}{\textbf{CV+}} uses \texttt{k}=10.}
    \label{fig:strategy_eval}
\end{figure}

Figure~\ref{fig:strategy_eval} compares recall and FDR across resampling-based strategies, which reuse the same $N$ reference observations for model fitting and calibration, and the split procedure, which divides the available data between training and calibration. It highlights how the effective calibration-set size determines the minimum attainable p-value and statistical power.

\paragraph{Probabilistic Approximation.}

With \texttt{Probabilistic()}, the package provides an approximate estimation that replaces empirical ranks with a kernel density estimates. This yields non-discrete and arbitrarily small \textit{p}-values, avoiding the empirical resolution floor. However, the variant is \underline{strictly not conformal}: finite-sample validity and FDR control are not guaranteed, and any asymptotic justification requires additional regularity assumptions.

\subsection{Marginal and Calibration-Conditional Error Control}

A useful distinction, already implicit in the discussion above, is that between \emph{marginal} and \emph{calibration-conditional} error control \citep{Bates2023}. Marginal error control averages over both the random calibration set and the random test sample. In this sense, it guarantees that conformal \textit{p}-values behave correctly across repeated applications of the full data-generating process. This is the standard form of validity attached to classical conformal methods and the one underlying the split-conformal guarantees introduced earlier.

By contrast, calibration-conditional control is stronger: it requires validity to hold after conditioning on the realized calibration set, rather than only on average over repeated calibration samples. This distinction is especially relevant in applications where one fixed reference dataset is collected once and then reused to screen many future observations.

In such settings, marginal validity may be formally correct yet still offer limited reassurance to the practitioner who must rely on one particular calibration set in deployment.

\begin{tcolorbox}[
  enhanced,
  listing engine=minted,
  minted language=bash,
  colback=orange!10,
  colframe=orange!60,
  coltitle=black,
  title=Implementation: Marginal Calibration,
  lower separated=true]
\begin{minted}{python}
detector = ConformalDetector(
    detector=IForest(),
    strategy=Split(1_000),
    estimation=Empirical()  # Controls marginally (default)
)
\end{minted}
\end{tcolorbox}

Operationally, the calibration-conditional approach starts from the same empirical conformal construction but then applies an additional simultaneous correction to the resulting \textit{p}-values. The parameter \texttt{delta} specifies the tolerated failure probability of this stronger guarantee: for example, \texttt{delta=0.1} means that, with probability at least \(0.9\) over the draw of the calibration set, the corrected \textit{p}-values are valid for that realized calibration sample. The argument \texttt{method="simes"} specifies the particular correction used to build this uniform bound, here a Simes-type adjustment following \citet{Bates2023}. Smaller values of \texttt{delta} give stronger assurance but typically produce more conservative \textit{p}-values.

\begin{tcolorbox}[
  enhanced,
  listing engine=minted,
  minted language=bash,
  colback=orange!10,
  colframe=orange!60,
  coltitle=black,
  title=Implementation: Calibration-conditional Calibration,
  lower separated=true]
\begin{minted}{python}
detector = ConformalDetector(
    detector=IForest(),
    strategy=Split(1_000),
    estimation=ConditionalEmpirical(method="simes", delta=0.1)
)
\end{minted}
\end{tcolorbox}

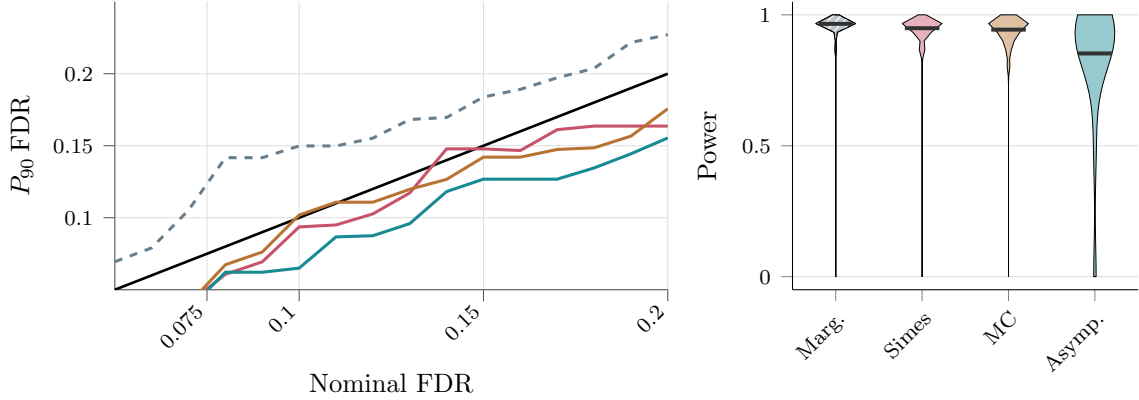
\begin{figure}[htbp]
    \centering

    \begin{tikzpicture}

    \pgfplotsset{
        violinEmp/.style={
            draw=black,
            very thin,
            line join=round,
            fill=evalteal!45,
            postaction={pattern=north east lines, pattern color=white}
        },
        violinSimes/.style={
            draw=black,
            very thin,
            line join=round,
            fill=evalmustard!45
        },
        violinMC/.style={
            draw=black,
            very thin,
            line join=round,
            fill=evalsky!45
        },
        violinAsymp/.style={
            draw=black,
            very thin,
            line join=round,
            fill=evalmagenta!45
        },
        meanbar/.style={
            black!80,
            line width=1.5pt,
            forget plot
        }
    }

    \begin{axis}[
        name=fdrplot,
        width=0.48\linewidth,
        height=0.25\linewidth,
        scale only axis,
        xmin=0.05, xmax=0.20,
        ymin=0.05, ymax=0.25,
        xlabel={Nominal FDR},
        ylabel={$P_{90}\,\mathrm{FDR}$},
        tick label style={
            font=\scriptsize,
            /pgf/number format/fixed,
            /pgf/number format/precision=3
        },
        x tick label style={rotate=45, anchor=east},
        label style={font=\small},
        xtick={0.075,0.10,0.15,0.20},
        ytick={0.10,0.15,0.20},
        grid=major,
        major grid style={draw=gray!25},
        axis x line*=bottom,
        axis y line*=left,
        tick align=outside,
        axis line style={draw=black!70},
        tick style={draw=black!70}
    ]

    \addplot[
        solid,
        black,
        line width=1.0pt,
        domain=0.05:0.20,
        samples=2
    ] {x};

    \addplot[color=evalteal, dashed, line width=1.15pt] coordinates {(0.050000,0.069457) (0.060000,0.079079) (0.070000,0.105295) (0.080000,0.141700) (0.090000,0.141700) (0.100000,0.149850) (0.110000,0.149850) (0.120000,0.155268) (0.130000,0.168210) (0.140000,0.169636) (0.150000,0.183852) (0.160000,0.189184) (0.170000,0.197210) (0.180000,0.203727) (0.190000,0.221681) (0.200000,0.227162)};

    \addplot[color=evalmustard, solid, line width=1.15pt] coordinates {(0.050000,0.033032) (0.060000,0.040301) (0.070000,0.040530) (0.080000,0.060731) (0.090000,0.069384) (0.100000,0.093636) (0.110000,0.094997) (0.120000,0.102676) (0.130000,0.117224) (0.140000,0.147826) (0.150000,0.147826) (0.160000,0.146679) (0.170000,0.161155) (0.180000,0.163654) (0.190000,0.163654) (0.200000,0.163654)};

    \addplot[color=evalsky, solid, line width=1.15pt] coordinates {(0.050000,0.021587) (0.060000,0.040000) (0.070000,0.040040) (0.080000,0.067373) (0.090000,0.076190) (0.100000,0.101852) (0.110000,0.110794) (0.120000,0.110794) (0.130000,0.119756) (0.140000,0.126659) (0.150000,0.142089) (0.160000,0.142089) (0.170000,0.147414) (0.180000,0.148561) (0.190000,0.156659) (0.200000,0.175682)};

    \addplot[color=evalmagenta, solid, line width=1.15pt] coordinates {(0.050000,0.023930) (0.060000,0.023930) (0.070000,0.036398) (0.080000,0.062165) (0.090000,0.062165) (0.100000,0.065009) (0.110000,0.086710) (0.120000,0.087519) (0.130000,0.095930) (0.140000,0.118182) (0.150000,0.126788) (0.160000,0.126788) (0.170000,0.126788) (0.180000,0.134571) (0.190000,0.144383) (0.200000,0.155411)};

    \end{axis}

    \begin{axis}[
        at={($(fdrplot.east)+(1.65cm,0)$)},
        anchor=west,
        width=0.3\linewidth,
        height=0.25\linewidth,
        scale only axis,
        xmin=0.5, xmax=4.5,
        ymin=-0.05, ymax=1.05,
        xtick={1,2,3,4},
        xticklabels={Marg., Simes, MC, Asymp.},
        xticklabel style={
            rotate=45,
            anchor=north east,
            font=\scriptsize,
            inner sep=2pt
        },
        yticklabel style={
            font=\scriptsize,
            text width=1.2em,
            align=right
        },
        tick label style={
            font=\scriptsize,
            /pgf/number format/fixed,
            /pgf/number format/precision=3
        },
        ylabel={Power},
        ylabel style={font=\small, align=center},
        ymajorgrids=true,
        grid style={gray!25},
        axis x line*=bottom,
        axis y line*=left,
        tick align=outside,
        clip=false
    ]

    \addplot[violinEmp] coordinates {(1.0000,0.0000) (1.0000,0.0333) (1.0000,0.0667) (1.0000,0.1000) (1.0000,0.1333) (1.0000,0.1667) (1.0000,0.2000) (1.0000,0.2333) (1.0000,0.2667) (1.0000,0.3000) (1.0000,0.3333) (1.0000,0.3667) (1.0000,0.4000) (1.0000,0.4333) (1.0000,0.4667) (1.0000,0.5000) (1.0000,0.5333) (1.0000,0.5667) (1.0000,0.6000) (1.0000,0.6333) (1.0000,0.6667) (1.0000,0.7000) (1.0000,0.7333) (1.0000,0.7667) (1.0000,0.8000) (1.0000,0.8333) (1.0088,0.8667) (1.0043,0.9000) (1.0201,0.9333) (1.2300,0.9667) (1.0369,1.0000) (0.9631,1.0000) (0.7700,0.9667) (0.9799,0.9333) (0.9957,0.9000) (0.9912,0.8667) (1.0000,0.8333) (1.0000,0.8000) (1.0000,0.7667) (1.0000,0.7333) (1.0000,0.7000) (1.0000,0.6667) (1.0000,0.6333) (1.0000,0.6000) (1.0000,0.5667) (1.0000,0.5333) (1.0000,0.5000) (1.0000,0.4667) (1.0000,0.4333) (1.0000,0.4000) (1.0000,0.3667) (1.0000,0.3333) (1.0000,0.3000) (1.0000,0.2667) (1.0000,0.2333) (1.0000,0.2000) (1.0000,0.1667) (1.0000,0.1333) (1.0000,0.1000) (1.0000,0.0667) (1.0000,0.0333) (1.0000,0.0000)} \closedcycle;

    \addplot[meanbar] coordinates {(0.8000,0.9655) (1.2000,0.9655)};

    \addplot[violinSimes] coordinates {(2.0000,0.0000) (2.0000,0.0333) (2.0000,0.0667) (2.0000,0.1000) (2.0000,0.1333) (2.0000,0.1667) (2.0000,0.2000) (2.0000,0.2333) (2.0000,0.2667) (2.0000,0.3000) (2.0000,0.3333) (2.0000,0.3667) (2.0000,0.4000) (2.0000,0.4333) (2.0000,0.4667) (2.0000,0.5000) (2.0000,0.5333) (2.0000,0.5667) (2.0000,0.6000) (2.0000,0.6333) (2.0000,0.6667) (2.0000,0.7000) (2.0000,0.7333) (2.0000,0.7667) (2.0001,0.8000) (2.0075,0.8333) (2.0324,0.8667) (2.0237,0.9000) (2.1066,0.9333) (2.2300,0.9667) (2.0442,1.0000) (1.9558,1.0000) (1.7700,0.9667) (1.8934,0.9333) (1.9763,0.9000) (1.9676,0.8667) (1.9925,0.8333) (1.9999,0.8000) (2.0000,0.7667) (2.0000,0.7333) (2.0000,0.7000) (2.0000,0.6667) (2.0000,0.6333) (2.0000,0.6000) (2.0000,0.5667) (2.0000,0.5333) (2.0000,0.5000) (2.0000,0.4667) (2.0000,0.4333) (2.0000,0.4000) (2.0000,0.3667) (2.0000,0.3333) (2.0000,0.3000) (2.0000,0.2667) (2.0000,0.2333) (2.0000,0.2000) (2.0000,0.1667) (2.0000,0.1333) (2.0000,0.1000) (2.0000,0.0667) (2.0000,0.0333) (2.0000,0.0000)} \closedcycle;

    \addplot[meanbar] coordinates {(1.8000,0.9490) (2.2000,0.9490)};

    \addplot[violinMC] coordinates {(3.0000,0.0000) (3.0000,0.0333) (3.0000,0.0667) (3.0000,0.1000) (3.0000,0.1333) (3.0000,0.1667) (3.0000,0.2000) (3.0000,0.2333) (3.0000,0.2667) (3.0000,0.3000) (3.0000,0.3333) (3.0000,0.3667) (3.0000,0.4000) (3.0000,0.4333) (3.0000,0.4667) (3.0000,0.5000) (3.0000,0.5333) (3.0000,0.5667) (3.0000,0.6000) (3.0000,0.6333) (3.0000,0.6667) (3.0000,0.7000) (3.0001,0.7333) (3.0031,0.7667) (3.0152,0.8000) (3.0138,0.8333) (3.0329,0.8667) (3.0656,0.9000) (3.1310,0.9333) (3.2300,0.9667) (3.0905,1.0000) (2.9095,1.0000) (2.7700,0.9667) (2.8690,0.9333) (2.9344,0.9000) (2.9671,0.8667) (2.9862,0.8333) (2.9848,0.8000) (2.9969,0.7667) (2.9999,0.7333) (3.0000,0.7000) (3.0000,0.6667) (3.0000,0.6333) (3.0000,0.6000) (3.0000,0.5667) (3.0000,0.5333) (3.0000,0.5000) (3.0000,0.4667) (3.0000,0.4333) (3.0000,0.4000) (3.0000,0.3667) (3.0000,0.3333) (3.0000,0.3000) (3.0000,0.2667) (3.0000,0.2333) (3.0000,0.2000) (3.0000,0.1667) (3.0000,0.1333) (3.0000,0.1000) (3.0000,0.0667) (3.0000,0.0333) (3.0000,0.0000)} \closedcycle;

    \addplot[meanbar] coordinates {(2.8000,0.9435) (3.2000,0.9435)};

    \addplot[violinAsymp] coordinates {(4.0140,0.0000) (4.0135,0.0333) (4.0122,0.0667) (4.0104,0.1000) (4.0083,0.1333) (4.0064,0.1667) (4.0050,0.2000) (4.0043,0.2333) (4.0045,0.2667) (4.0054,0.3000) (4.0070,0.3333) (4.0092,0.3667) (4.0115,0.4000) (4.0138,0.4333) (4.0159,0.4667) (4.0179,0.5000) (4.0200,0.5333) (4.0232,0.5667) (4.0282,0.6000) (4.0365,0.6333) (4.0492,0.6667) (4.0672,0.7000) (4.0908,0.7333) (4.1194,0.7667) (4.1509,0.8000) (4.1818,0.8333) (4.2081,0.8667) (4.2252,0.9000) (4.2300,0.9333) (4.2213,0.9667) (4.2003,1.0000) (3.7997,1.0000) (3.7787,0.9667) (3.7700,0.9333) (3.7748,0.9000) (3.7919,0.8667) (3.8182,0.8333) (3.8491,0.8000) (3.8806,0.7667) (3.9092,0.7333) (3.9328,0.7000) (3.9508,0.6667) (3.9635,0.6333) (3.9718,0.6000) (3.9768,0.5667) (3.9800,0.5333) (3.9821,0.5000) (3.9841,0.4667) (3.9862,0.4333) (3.9885,0.4000) (3.9908,0.3667) (3.9930,0.3333) (3.9946,0.3000) (3.9955,0.2667) (3.9957,0.2333) (3.9950,0.2000) (3.9936,0.1667) (3.9917,0.1333) (3.9896,0.1000) (3.9878,0.0667) (3.9865,0.0333) (3.9860,0.0000)} \closedcycle;

    \addplot[meanbar] coordinates {(3.8000,0.8525) (4.2000,0.8525)};

    \end{axis}

    \end{tikzpicture}

    \caption{
        \textbf{Conditional-calibration corrections reduce upper-tail empirical FDR at a cost in power.}
        Left: 90th-percentile empirical FDR across 20 randomized trials as a function of the nominal FDR target, illustrating tail behavior under conditional calibration with \texttt{delta=0.1}, with marginal control as reference.
        Right: distribution of recall across trials at nominal level $\alpha = 0.1$.
        Methods correspond to JaB+ with marginal control and calibration-conditional control via
        \texttt{method=}\textcolor{evalmustard}{\texttt{"simes"}},
        \textcolor{evalsky}{\texttt{"mc"}},
        and \textcolor{evalmagenta}{\texttt{"asymptotic"}}.
    }
    \label{fig:conditional}
\end{figure}

The trade-off is familiar: stronger conditioning generally requires more conservative procedures. Calibration-conditional guarantees therefore tend to come at some cost in power, since they adjust the empirical \textit{p}-values to remain reliable for the realized calibration sample rather than only on average across repeated samples. For CAD, this distinction becomes especially important once one moves beyond single-observation decisions and begins to study repeated or large-scale use of the same calibrated system. Figure~\ref{fig:conditional} compares marginal and calibration-conditional error control and its impact on testing power. 

%

\subsection{Assumptions for Standard Conformal Validity}
\label{sec:assumptions}

In the standard, unweighted setting, conformal validity rests on exchangeability between the calibration observations and the test observation under the null hypothesis. In most applications, this is ensured by assuming that these observations are sampled i.i.d.\ from a common inlier distribution, although exchangeability is formally slightly weaker than full independence and identical distribution. What matters is that, under the null, the calibration scores and the test score are generated symmetrically, so that none occupies a privileged role in the conformal comparison.

A second requirement concerns the separation between model fitting and
calibration. The anomaly detector may otherwise be arbitrary, but after it is trained on \(\mathcal{D}_{\mathrm{train}}\), including any independent algorithmic randomness, the resulting scoring rule must remain fixed while calibration and test observations are scored; it must not adapt to either set. This is why validity is largely model agnostic. Nevertheless, the detector remains crucial for statistical power: a weak or unstable scoring rule may still yield valid \textit{p}-values, but they can be uninformative and lead to few true discoveries.

Under these conditions, the resulting conformal
\textit{p}-values are marginally valid. That is, their type-I error guarantee holds after averaging over both the random calibration sample and the random test point. As discussed above, stronger calibration-conditional guarantees are possible, but they require additional adjustment and typically come with a corresponding loss in power. The classical exchangeability framework is therefore the natural starting point for CAD. The next section considers how this picture changes once exchangeability is no longer tenable (due to covariate/feature shift) and the calibration and test distributions must be related through a more general weighting scheme.

\subsection{Weighted Conformal Anomaly Detection}
\label{sec:wcad}

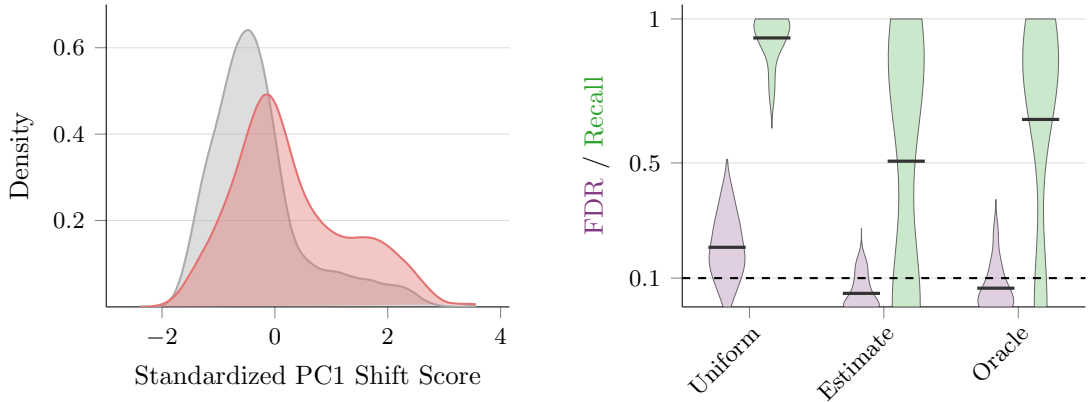
\begin{figure}[htbp]
    \centering
    \begin{tikzpicture}[
        fdrStyle/.style={
            draw=black,
            very thin,
            line join=round,
            fill=wgold!40,
            opacity=0.6
        },
        powerStyle/.style={
            draw=black,
            very thin,
            line join=round,
            fill=wforest!40,
            opacity=0.6
        },
        meanmarker/.style={
            only marks,
            mark=-,
            mark size=7pt,
            line width=1.2pt,
            color=black!80
        }
    ]
        \begin{groupplot}[
            group style={
                group size=2 by 1,
                horizontal sep=0.15\linewidth,
            },
            width=0.35\linewidth,
            height=4cm,
            scale only axis,
            ymajorgrids=true,
            grid style={gray!25},
            axis x line*=bottom,
            axis y line*=left,
            tick align=outside,
            tick label style={font=\footnotesize},
            label style={font=\small}
        ]
        
        \nextgroupplot[
            xlabel={Standardized PC1 Shift Score},
            ylabel={Density},
            ymin=0, ymax=0.7,
            ytick={0.2,0.4,0.6},
            legend pos=north east,
            legend style={draw=none, fill=none, font=\footnotesize} 
        ]
            \addplot[
                wslate!65, thick, smooth, fill=wslate!65, fill opacity=0.4
            ] coordinates {(-2.393,0.001) (-2.271,0.001) (-2.150,0.001) (-2.028,0.002) (-1.907,0.010) (-1.785,0.030) (-1.664,0.071) (-1.542,0.134) (-1.421,0.211) (-1.299,0.289) (-1.178,0.355) (-1.056,0.414) (-0.935,0.478) (-0.813,0.545) (-0.692,0.600) (-0.570,0.633) (-0.449,0.640) (-0.327,0.614) (-0.206,0.554) (-0.084,0.467) (0.037,0.367) (0.159,0.266) (0.280,0.187) (0.402,0.140) (0.523,0.116) (0.645,0.101) (0.766,0.090) (0.888,0.085) (1.009,0.083) (1.131,0.080) (1.252,0.075) (1.374,0.070) (1.495,0.066) (1.617,0.064) (1.738,0.061) (1.860,0.055) (1.981,0.052) (2.103,0.050) (2.224,0.048) (2.346,0.042) (2.467,0.033) (2.589,0.021) (2.710,0.012) (2.832,0.006) (2.953,0.003) (3.075,0.002) (3.196,0.001) (3.318,0.001) (3.439,0.002) (3.561,0.002)};
            
            \addplot[
                wrose!65, thick, smooth, fill=wrose!65, fill opacity=0.4
            ] coordinates {(-2.393,0.000) (-2.271,0.000) (-2.150,0.002) (-2.028,0.005) (-1.907,0.012) (-1.785,0.024) (-1.664,0.043) (-1.542,0.068) (-1.421,0.095) (-1.299,0.124) (-1.178,0.153) (-1.056,0.187) (-0.935,0.224) (-0.813,0.266) (-0.692,0.313) (-0.570,0.365) (-0.449,0.417) (-0.327,0.462) (-0.206,0.489) (-0.084,0.489) (0.037,0.462) (0.159,0.414) (0.280,0.358) (0.402,0.304) (0.523,0.261) (0.645,0.229) (0.766,0.206) (0.888,0.188) (1.009,0.175) (1.131,0.166) (1.252,0.161) (1.374,0.160) (1.495,0.161) (1.617,0.161) (1.738,0.159) (1.860,0.153) (1.981,0.143) (2.103,0.130) (2.224,0.115) (2.346,0.100) (2.467,0.082) (2.589,0.063) (2.710,0.045) (2.832,0.029) (2.953,0.017) (3.075,0.010) (3.196,0.008) (3.318,0.007) (3.439,0.007) (3.561,0.006)};

        \nextgroupplot[
            ymin=0, ymax=1.05,
            xmin=0, xmax=9,
            xtick={1.5, 4.5, 7.5},
            ytick={0.1, 0.5, 1.0},
            ylabel={\textcolor{wgold}{FDR} / \textcolor{wforest}{Recall}},
            xticklabels={Uniform, {Estimate}, Oracle},
            xticklabel style={rotate=45, anchor=north east, font=\footnotesize, inner sep=1pt, align=center}
        ]

            \addplot[fdrStyle] coordinates {(1.089,0.000) (1.109,0.010) (1.133,0.021) (1.156,0.031) (1.183,0.042) (1.207,0.052) (1.233,0.063) (1.257,0.073) (1.283,0.084) (1.305,0.094) (1.326,0.104) (1.348,0.115) (1.365,0.125) (1.381,0.136) (1.391,0.146) (1.398,0.157) (1.400,0.167) (1.398,0.178) (1.393,0.188) (1.385,0.198) (1.374,0.209) (1.364,0.219) (1.352,0.230) (1.342,0.240) (1.330,0.251) (1.320,0.261) (1.308,0.272) (1.297,0.282) (1.284,0.292) (1.270,0.303) (1.256,0.313) (1.240,0.324) (1.225,0.334) (1.210,0.345) (1.195,0.355) (1.180,0.366) (1.166,0.376) (1.153,0.386) (1.138,0.397) (1.124,0.407) (1.108,0.418) (1.094,0.428) (1.078,0.439) (1.065,0.449) (1.052,0.460) (1.041,0.470) (1.031,0.480) (1.023,0.491) (1.016,0.501) (1.011,0.512) (0.989,0.512) (0.984,0.501) (0.977,0.491) (0.969,0.480) (0.959,0.470) (0.948,0.460) (0.935,0.449) (0.922,0.439) (0.906,0.428) (0.892,0.418) (0.876,0.407) (0.862,0.397) (0.847,0.386) (0.834,0.376) (0.820,0.366) (0.805,0.355) (0.790,0.345) (0.775,0.334) (0.760,0.324) (0.744,0.313) (0.730,0.303) (0.716,0.292) (0.703,0.282) (0.692,0.272) (0.680,0.261) (0.670,0.251) (0.658,0.240) (0.648,0.230) (0.636,0.219) (0.626,0.209) (0.615,0.198) (0.607,0.188) (0.602,0.178) (0.600,0.167) (0.602,0.157) (0.609,0.146) (0.619,0.136) (0.635,0.125) (0.652,0.115) (0.674,0.104) (0.695,0.094) (0.717,0.084) (0.743,0.073) (0.767,0.063) (0.793,0.052) (0.817,0.042) (0.844,0.031) (0.867,0.021) (0.891,0.010) (0.911,0.000)};
            \addplot+[meanmarker, forget plot] coordinates {(1.000,0.207)};
            
            \addplot[powerStyle] coordinates {(2.002,0.620) (2.003,0.628) (2.004,0.636) (2.006,0.643) (2.008,0.651) (2.011,0.659) (2.015,0.667) (2.019,0.674) (2.023,0.682) (2.029,0.690) (2.034,0.698) (2.039,0.705) (2.045,0.713) (2.050,0.721) (2.054,0.729) (2.058,0.736) (2.062,0.744) (2.064,0.752) (2.067,0.760) (2.068,0.767) (2.069,0.775) (2.071,0.783) (2.072,0.791) (2.073,0.798) (2.075,0.806) (2.077,0.814) (2.081,0.822) (2.085,0.829) (2.092,0.837) (2.101,0.845) (2.113,0.853) (2.126,0.860) (2.143,0.868) (2.164,0.876) (2.186,0.884) (2.208,0.891) (2.233,0.899) (2.258,0.907) (2.283,0.915) (2.304,0.922) (2.326,0.930) (2.346,0.938) (2.364,0.946) (2.378,0.953) (2.390,0.961) (2.398,0.969) (2.400,0.977) (2.396,0.984) (2.386,0.992) (2.367,1.000) (1.633,1.000) (1.614,0.992) (1.604,0.984) (1.600,0.977) (1.602,0.969) (1.610,0.961) (1.622,0.953) (1.636,0.946) (1.654,0.938) (1.674,0.930) (1.696,0.922) (1.717,0.915) (1.742,0.907) (1.767,0.899) (1.792,0.891) (1.814,0.884) (1.836,0.876) (1.857,0.868) (1.874,0.860) (1.887,0.853) (1.899,0.845) (1.908,0.837) (1.915,0.829) (1.919,0.822) (1.923,0.814) (1.925,0.806) (1.927,0.798) (1.928,0.791) (1.929,0.783) (1.931,0.775) (1.932,0.767) (1.933,0.760) (1.936,0.752) (1.938,0.744) (1.942,0.736) (1.946,0.729) (1.950,0.721) (1.955,0.713) (1.961,0.705) (1.966,0.698) (1.971,0.690) (1.977,0.682) (1.981,0.674) (1.985,0.667) (1.989,0.659) (1.992,0.651) (1.994,0.643) (1.996,0.636) (1.997,0.628) (1.998,0.620)};
            \addplot+[meanmarker, forget plot] coordinates {(2.000,0.934)};

            \addplot[fdrStyle] coordinates {(4.395,0.000) (4.400,0.006) (4.396,0.011) (4.382,0.017) (4.365,0.022) (4.341,0.028) (4.320,0.033) (4.294,0.039) (4.274,0.044) (4.252,0.050) (4.233,0.056) (4.220,0.061) (4.206,0.067) (4.197,0.072) (4.186,0.078) (4.179,0.083) (4.171,0.089) (4.164,0.095) (4.159,0.100) (4.154,0.106) (4.152,0.111) (4.149,0.117) (4.147,0.122) (4.145,0.128) (4.143,0.133) (4.139,0.139) (4.134,0.145) (4.128,0.150) (4.119,0.156) (4.111,0.161) (4.099,0.167) (4.089,0.172) (4.076,0.178) (4.066,0.183) (4.054,0.189) (4.043,0.195) (4.035,0.200) (4.026,0.206) (4.020,0.211) (4.015,0.217) (4.011,0.222) (4.007,0.228) (4.005,0.233) (4.003,0.239) (4.002,0.245) (4.001,0.250) (4.001,0.256) (4.000,0.261) (4.000,0.267) (4.000,0.272) (4.000,0.272) (4.000,0.267) (4.000,0.261) (3.999,0.256) (3.999,0.250) (3.998,0.245) (3.997,0.239) (3.995,0.233) (3.993,0.228) (3.989,0.222) (3.985,0.217) (3.980,0.211) (3.974,0.206) (3.965,0.200) (3.957,0.195) (3.946,0.189) (3.934,0.183) (3.924,0.178) (3.911,0.172) (3.901,0.167) (3.889,0.161) (3.881,0.156) (3.872,0.150) (3.866,0.145) (3.861,0.139) (3.857,0.133) (3.855,0.128) (3.853,0.122) (3.851,0.117) (3.848,0.111) (3.846,0.106) (3.841,0.100) (3.836,0.095) (3.829,0.089) (3.821,0.083) (3.814,0.078) (3.803,0.072) (3.794,0.067) (3.780,0.061) (3.767,0.056) (3.748,0.050) (3.726,0.044) (3.706,0.039) (3.680,0.033) (3.659,0.028) (3.635,0.022) (3.618,0.017) (3.604,0.011) (3.600,0.006) (3.605,0.000)};
            \addplot+[meanmarker, forget plot] coordinates {(4.000,0.047)};
            
            \addplot[powerStyle] coordinates {(5.313,0.000) (5.312,0.020) (5.310,0.041) (5.305,0.061) (5.298,0.082) (5.290,0.102) (5.280,0.122) (5.269,0.143) (5.257,0.163) (5.245,0.184) (5.233,0.204) (5.221,0.224) (5.209,0.245) (5.199,0.265) (5.189,0.286) (5.181,0.306) (5.175,0.327) (5.170,0.347) (5.167,0.367) (5.167,0.388) (5.168,0.408) (5.171,0.429) (5.176,0.449) (5.183,0.469) (5.192,0.490) (5.202,0.510) (5.214,0.531) (5.226,0.551) (5.240,0.571) (5.255,0.592) (5.270,0.612) (5.286,0.633) (5.301,0.653) (5.316,0.673) (5.331,0.694) (5.345,0.714) (5.358,0.735) (5.369,0.755) (5.380,0.776) (5.388,0.796) (5.394,0.816) (5.398,0.837) (5.400,0.857) (5.400,0.878) (5.397,0.898) (5.392,0.918) (5.384,0.939) (5.374,0.959) (5.362,0.980) (5.349,1.000) (4.651,1.000) (4.638,0.980) (4.626,0.959) (4.616,0.939) (4.608,0.918) (4.603,0.898) (4.600,0.878) (4.600,0.857) (4.602,0.837) (4.606,0.816) (4.612,0.796) (4.620,0.776) (4.631,0.755) (4.642,0.735) (4.655,0.714) (4.669,0.694) (4.684,0.673) (4.699,0.653) (4.714,0.633) (4.730,0.612) (4.745,0.592) (4.760,0.571) (4.774,0.551) (4.786,0.531) (4.798,0.510) (4.808,0.490) (4.817,0.469) (4.824,0.449) (4.829,0.429) (4.832,0.408) (4.833,0.388) (4.833,0.367) (4.830,0.347) (4.825,0.327) (4.819,0.306) (4.811,0.286) (4.801,0.265) (4.791,0.245) (4.779,0.224) (4.767,0.204) (4.755,0.184) (4.743,0.163) (4.731,0.143) (4.720,0.122) (4.710,0.102) (4.702,0.082) (4.695,0.061) (4.690,0.041) (4.688,0.020) (4.687,0.000)};
            \addplot+[meanmarker, forget plot] coordinates {(5.000,0.506)};

            \addplot[fdrStyle] coordinates {(7.370,0.000) (7.388,0.008) (7.398,0.015) (7.400,0.023) (7.396,0.030) (7.384,0.038) (7.367,0.046) (7.349,0.053) (7.327,0.061) (7.305,0.068) (7.281,0.076) (7.258,0.084) (7.238,0.091) (7.218,0.099) (7.202,0.106) (7.186,0.114) (7.172,0.122) (7.161,0.129) (7.150,0.137) (7.139,0.145) (7.130,0.152) (7.119,0.160) (7.110,0.167) (7.100,0.175) (7.090,0.183) (7.083,0.190) (7.075,0.198) (7.070,0.205) (7.066,0.213) (7.063,0.221) (7.061,0.228) (7.059,0.236) (7.058,0.243) (7.057,0.251) (7.055,0.259) (7.053,0.266) (7.049,0.274) (7.045,0.281) (7.040,0.289) (7.034,0.297) (7.029,0.304) (7.024,0.312) (7.019,0.319) (7.015,0.327) (7.011,0.335) (7.008,0.342) (7.006,0.350) (7.004,0.358) (7.003,0.365) (7.002,0.373) (6.998,0.373) (6.997,0.365) (6.996,0.358) (6.994,0.350) (6.992,0.342) (6.989,0.335) (6.985,0.327) (6.981,0.319) (6.976,0.312) (6.971,0.304) (6.966,0.297) (6.960,0.289) (6.955,0.281) (6.951,0.274) (6.947,0.266) (6.945,0.259) (6.943,0.251) (6.942,0.243) (6.941,0.236) (6.939,0.228) (6.937,0.221) (6.934,0.213) (6.930,0.205) (6.925,0.198) (6.917,0.190) (6.910,0.183) (6.900,0.175) (6.890,0.167) (6.881,0.160) (6.870,0.152) (6.861,0.145) (6.850,0.137) (6.839,0.129) (6.828,0.122) (6.814,0.114) (6.798,0.106) (6.782,0.099) (6.762,0.091) (6.742,0.084) (6.719,0.076) (6.695,0.068) (6.673,0.061) (6.651,0.053) (6.633,0.046) (6.616,0.038) (6.604,0.030) (6.600,0.023) (6.602,0.015) (6.612,0.008) (6.630,0.000)};
            \addplot+[meanmarker, forget plot] coordinates {(7.000,0.065)};
            
            \addplot[powerStyle] coordinates {(8.143,0.000) (8.142,0.020) (8.140,0.041) (8.137,0.061) (8.132,0.082) (8.127,0.102) (8.121,0.122) (8.113,0.143) (8.106,0.163) (8.099,0.184) (8.092,0.204) (8.085,0.224) (8.079,0.245) (8.073,0.265) (8.069,0.286) (8.066,0.306) (8.065,0.327) (8.065,0.347) (8.066,0.367) (8.070,0.388) (8.075,0.408) (8.083,0.429) (8.092,0.449) (8.103,0.469) (8.117,0.490) (8.131,0.510) (8.148,0.531) (8.166,0.551) (8.184,0.571) (8.205,0.592) (8.225,0.612) (8.247,0.633) (8.268,0.653) (8.289,0.673) (8.309,0.694) (8.328,0.714) (8.346,0.735) (8.361,0.755) (8.375,0.776) (8.385,0.796) (8.393,0.816) (8.398,0.837) (8.400,0.857) (8.399,0.878) (8.394,0.898) (8.387,0.918) (8.376,0.939) (8.364,0.959) (8.348,0.980) (8.331,1.000) (7.669,1.000) (7.652,0.980) (7.636,0.959) (7.624,0.939) (7.613,0.918) (7.606,0.898) (7.601,0.878) (7.600,0.857) (7.602,0.837) (7.607,0.816) (7.615,0.796) (7.625,0.776) (7.639,0.755) (7.654,0.735) (7.672,0.714) (7.691,0.694) (7.711,0.673) (7.732,0.653) (7.753,0.633) (7.775,0.612) (7.795,0.592) (7.816,0.571) (7.834,0.551) (7.852,0.531) (7.869,0.510) (7.883,0.490) (7.897,0.469) (7.908,0.449) (7.917,0.429) (7.925,0.408) (7.930,0.388) (7.934,0.367) (7.935,0.347) (7.935,0.327) (7.934,0.306) (7.931,0.286) (7.927,0.265) (7.921,0.245) (7.915,0.224) (7.908,0.204) (7.901,0.184) (7.894,0.163) (7.887,0.143) (7.879,0.122) (7.873,0.102) (7.868,0.082) (7.863,0.061) (7.860,0.041) (7.858,0.020) (7.857,0.000)};
            \addplot+[meanmarker, forget plot] coordinates {(8.000,0.651)};
            \draw[black, dashed, thick] (axis cs:0,0.1) -- (axis cs:9,0.1) node[pos=0.03, above right, font=\footnotesize]{};
        \end{groupplot}
    \end{tikzpicture}
    \caption{\textbf{Importance weighting restores empirical marginal FDR control in this covariate shift scenario.\protect\footnotemark}
Left: density of the standardized first principal-component shift score demonstrating covariate shift between the calibration and \textcolor{wrose}{test distribution}. Right: distribution of empirical marginal FDR and recall at nominal level $\alpha = 0.1$ for uniform weighting, estimated weighting (logistic), and oracle weighting with weighted conformal selection.}
\label{fig:weighted}
\end{figure}

The standard conformal construction relies on exchangeability between calibration and test observations. This assumption can be too restrictive when the notion of normality remains unchanged, but the distribution of inlier covariates shifts between calibration and deployment. In this case, the calibration sample is no longer representative of the test environment, and conformal \textit{p}-values may lose their validity, see Figure \ref{fig:weighted}.

Weighted CAD addresses this covariate-shift setting by replacing the unweighted empirical comparison with a weighted one. Let calibration inliers be sampled from a source distribution \(P\), while test inliers follow a target distribution \(Q\). The shift is assumed to be captured by a covariate-dependent likelihood ratio (i.e. a \textit{Radon--Nikod\'ym derivative})
\[
w(x) = \frac{\mathrm{d}Q_X}{\mathrm{d}P_X}(x),
\]
or a suitable estimate thereof \citep{Shimodaira2000}, for example via probabilistic Random Forest classification. A support condition is required: \(Q_X\) must be absolutely continuous with respect to \(P_X\) with $\text{supp}(Q_X) \subseteq \text{supp}(P_X)$, so that covariate values occurring at test time have positive probability under the calibration distribution. Regions outside calibration support cannot be empirically corrected by reweighting.

Using the weights \(w(X_i)\), calibration scores that are more representative of the target environment contribute more strongly to the conformal reference distribution, while less representative scores contribute less. The resulting weighted conformal \textit{p}-values therefore compare a test observation to a target-adjusted inlier reference distribution rather than to the source calibration sample alone, so rank counts are replaced by weighted rank counts.

\footnotetext{Covariate shift induced by rejection-sampling along 1st inlier principal component \citet{Jin2025}.}

\subsection{Weighted Conformal Validity and Multiple Testing}
\label{sec:assumptions_weighted}

Weighted conformal validity replaces ordinary exchangeability with weighted exchangeability induced by the covariate-shift assumption. Calibration and test observations may have different covariate distributions, but their difference must be captured by the density ratio \(w(x)\), while the relevant conditional notion of normality given \(X\) remains stable. Thus, weighted conformal inference addresses covariate shift with overlap, not arbitrary concept shifts between disconnected distributions.

Under these assumptions, and with the correct weights, weighted conformal \textit{p}-values are marginally valid for a single test under the target distribution: thresholding one such \textit{p}-value at level \(\alpha\) controls the type-I error at level \(\alpha\). If \(w\) is unknown, it is commonly replaced by a density-ratio estimate learned from calibration and target covariates, but exact validity is then no longer automatic and depends on the quality of this estimate. Practically, this is why sufficient overlap is important. If the target distribution concentrates in regions poorly represented by calibration data, a few observations receive large weights and the calibration may become unstable or ineffective for discovery \citep{Hennhofer2026}.

This single-test validity does not automatically imply finite-sample multiple-testing guarantees. In the unweighted setting, the shared calibration sample induces dependence across test \textit{p}-values, but this dependence is structured enough to support e.g. standard BH. Under weighting, the resulting \textit{p}-values need not satisfy positive regression dependence on a subset (PRDS), the positive-dependence property used in the usual BH argument, because the covariate-dependent weights also enter the shared calibration comparison.

Consequently, BH may remain useful in practice, and FDR control may hold asymptotically, but exact finite-sample FDR control generally requires procedures tailored to the weighted setting, such as Weighted Conformalized Selection (WCS) \citep{Jin2025}. Weighted CAD should therefore be viewed not only as an alternative calibration formula, but also as a shift-aware extension that changes the downstream multiple-testing problem.

\begin{tcolorbox}[
  enhanced,
  listing engine=minted,
  minted language=bash,
  colback=orange!10,
  colframe=orange!60,
  coltitle=black,
  title=Implementation: Weighted Conformal Anomaly Detection,
  lower separated=true]
\begin{minted}{python}
detector = ConformalDetector(
    detector=IForest(),
    strategy=Split(1_000),
    weight_estimator=forest_weight_estimator()
).fit(x_train)

# Dispatches to WCS if a 'weight_estimator' (above) is defined
decisions = detector.select(
    x_test,
    alpha=0.1
)
\end{minted}
\end{tcolorbox}

\section{Conformal Martingales and Change-Point Detection}
\label{sec:martingale}

The CAD framework developed so far is primarily pointwise: each new observation receives a conformal \textit{p}-value and is assessed either individually or as part of a multiple-testing problem. In sequential settings, however, the central question is often different. Rather than asking whether a single observation is anomalous, one may want to detect whether the data-generating mechanism itself has changed over time. This is where \textit{conformal martingales} become natural. Within anomaly detection, they are best viewed not as a replacement for batch outlier screening, but as a process-level method for temporal anomaly detection, with change-point detection as the main use case.

The basic idea is to monitor a stream of conformal \textit{p}-values over time. Under the null hypothesis of exchangeability, these \textit{p}-values are valid; in the ideal randomized online conformal construction, they behave like independent draws from the uniform distribution on $[0,1]$. A conformal martingale transforms the sequence $p_1, p_2,\dots$ into a nonnegative evidence process $M_0, M_1, M_2, \dots$
that is fair, in a betting sense, under the null hypothesis. Consequently, if $M_n$ grows to a large value, this indicates accumulated evidence against the assumption that the stream is still behaving as before. A common way to turn this evidence process into an alarm is to use a \textit{Ville threshold}: one raises an alarm once $M_n$ crosses a fixed level $\lambda$. For a nonnegative martingale started at one, Ville's inequality gives the anytime bound
\[
\mathbb{P}\!\left(\sup_{t \geq 0} M_t \geq \lambda\right) \leq \frac{1}{\lambda},
\]
so the probability of crossing the threshold under the null is at most $1/\lambda$. For example, $\lambda=100$ corresponds to an anytime false-alarm bound of at most $1\%$ for a monitored stream.

\begin{tcolorbox}[
  breakable,
  enhanced,
  listing engine=minted,
  minted language=bash,
  colback=orange!10,
  colframe=orange!60,
  coltitle=black,
  title=Implementation: Exchangeability Martingale,
  lower separated=true]
\begin{minted}{python}
from sklearn.ensemble import IsolationForest

from nonconform import ConformalDetector, Split
from nonconform.martingales import AlarmConfig, PowerMartingale

detector = ConformalDetector(
    detector=IsolationForest(random_state=42),
    strategy=Split(n_calib=0.2)
).fit(x_train)

martingale = PowerMartingale(
    epsilon=0.5,
    alarm_config=AlarmConfig(ville_threshold=100)
)

for x_t in data_stream:
    p_t = detector.compute_p_value(x_t)
    state = martingale.update(p_t)

    if "ville" in state.triggered_alarms:
        print("Alert")
\end{minted}
\end{tcolorbox}

A classical example is the power martingale \citep{Vovk2003}
\[
M_n^{(\varepsilon)}
=
\prod_{i=1}^n \varepsilon p_i^{\varepsilon-1},
\qquad \varepsilon \in (0,1],
\]
which grows when unusually many small conformal \textit{p}-values are observed. Since the performance of a fixed value of $\varepsilon$ can depend strongly on the form of the departure from exchangeability, one often works instead with mixtures over different betting parameters \citep{Vovk2003}. Adaptive plug-in variants estimate the betting strategy from previous \textit{p}-values \citep{Fedorova2012}. In this way, conformal martingales provide an online mechanism for accumulating evidence against the null, rather than issuing single-step decisions.

This perspective connects naturally to change-point detection. An isolated small \textit{p}-value may correspond only to a transient or local anomaly. By contrast, a sustained run of smaller \textit{p}-values leads to systematic martingale growth and is therefore more indicative of a structural shift in the underlying process. Conformal martingales are useful when the goal is to determine whether a previously calibrated detector or predictive model is no longer operating under the assumptions on which its validity was based. They therefore provide a statistically principled bridge between CAD and online change-point detection.

The current \texttt{nonconform} implementation follows this interpretation directly: it operates on sequential conformal \textit{p}-values rather than on raw anomaly scores, and provides power, simple-mixture, and simple-jumper martingales. In addition to the cumulative martingale evidence process, the implementation supports Ville thresholding of the standard martingale and a restart-mixture e-process, which improves sensitivity to later changes while preserving anytime false-alarm control when the input sequence is
conditionally valid. The package deliberately keeps downstream decision logic outside the martingale classes: a martingale alarm signals accumulated evidence of distributional change, while the response---such as retraining, or recalibration---remains a separate design choice \citep{Vovk2021b}.


\begin{figure}[H]
    \centering
    \begin{tikzpicture}[
        spy using outlines={rectangle, 
            magnification=2, 
            width=3cm,              
            height=2cm,             
            connect spies,
            every spy on node/.append style={thick},
            every spy in node/.append style={thick}
        }
    ]

        \begin{axis}[
            width=13.15cm,
            height=5cm,
            scale only axis,
            xmin=0, xmax=2000,
            ymin=0, ymax=1.05,
            axis lines=none, 
            axis on top
        ]
            \shade [left color=white, right color=red!15]
                (axis cs:1000, 0) rectangle (axis cs:2000, 1.05);

            \addplot [black, very thin, forget plot]
                coordinates {(1000, 0) (1000, 1.05)};
        \end{axis}

        \begin{axis}[
            name=main_plot,
            width=13.15cm,
            height=5cm,
            scale only axis,
            ymode=log,
            xmin=0, xmax=2000,
            ymin=1e-100, ymax=1e220,
            axis y line*=left,
            axis x line*=bottom,
            ylabel={Martingale Value},
            ylabel near ticks,
            xtick={500, 1000, 1500},
        ]

            \addplot [
                color=wforest,
                very thick,
                mark=none,
            ] table [
                x expr=\coordindex+1,
                y=martingale,
                col sep=comma
            ] {figures/4-martingale-plots/exchangeability_martingale_showcase_stream.csv};

            \addplot [
                color=wforestPlum,
                very thick,
                mark=none,
            ] table [
                x expr=\coordindex+1,
                y=restarted_martingale,
                col sep=comma
            ] {figures/4-martingale-plots/exchangeability_martingale_showcase_stream.csv};

            \addplot [wforest, thick, dashed, forget plot]
                coordinates {(1481, 592.07) (1481, 1e170)};
            \node [circle, fill=white, draw=red, line width=0.6pt, inner sep=0pt, minimum size=12pt, font=\bfseries, text=red] 
                at (axis cs:1481, 1e170) {!};

            \addplot [wforestPlum, thick, dashed, forget plot]
                coordinates {(1242, 452.54) (1242, 1e160)};
            \node [circle, fill=white, draw=red, line width=0.6pt, inner sep=0pt, minimum size=12pt, font=\bfseries, text=red] 
                at (axis cs:1242, 1e170) {!};

            \addplot [dashed, darkgray, thick, forget plot] 
                table [x expr=\coordindex+1, y=ville_threshold, col sep=comma] 
                {figures/4-martingale-plots/exchangeability_martingale_showcase_stream.csv};
            
            \addplot [dashed, orange, thick, forget plot] 
                table [x expr=\coordindex+1, y=restarted_ville_threshold, col sep=comma] 
                {figures/4-martingale-plots/exchangeability_martingale_showcase_stream.csv};

            \coordinate (first_crossing) at (axis cs:1242, 1000);
            
            \coordinate (lens_pos) at (axis cs:500, 1e110);

        \end{axis}

        \spy [black!60] on (first_crossing) in node at (lens_pos);

    \end{tikzpicture}
    \caption{\textbf{Exchangeability martingales detect distributional changes in data streams.} A stream of 2,000 observations is processed via Isolation Forest, producing $p$-values that are approximately uniform before the change point and become non-uniform as the anomaly rate increases linearly from 0\% to 100\% after $t=1000$. The \textcolor{wforest}{\textbf{Standard Martingale}} and \textcolor{wforestPlum}{\textbf{Restart-mixture Martingale}} evaluate $p$-values sequentially and trigger alarms when crossing the threshold $1/\alpha$, with $\alpha=0.001$. For each Ville-valid statistic, this controls the anytime false-alarm probability at no more than 0.1\% under an exchangeability null. The stated Ville bound applies only when sequential \textit{p}-values generate conditionally e-valid betting factors. The fixed-split paths shown here are illustrative processes.}
\end{figure}
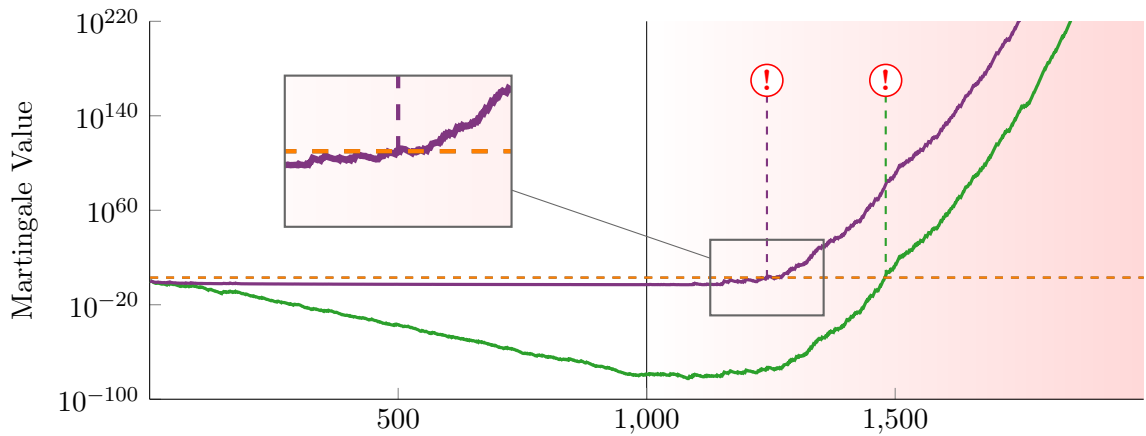

Conformal martingales should be understood as a complementary extension of CAD. Standard conformal methods answer the question \enquote{\textit{which observations look unusual relative to a reference sample?}} Conformal martingales instead address the sequential question \enquote{\textit{is the stream, as a process, still behaving as if it were generated under the same exchangeable regime?}} This makes them particularly useful in monitoring and deployment settings, where the onset of a new regime may matter more than any single anomalous point observation.

\section{The \texttt{nonconform} Package}
\label{sec:software}

CAD in practice requires more than a scoring model: it also needs calibrated score routing, valid $p$-value construction, and multiplicity-aware decision rules. While Python provides mature conformal prediction libraries, including \texttt{MAPIE} \citep{Cordier2023}, \texttt{crepes} \citep{Bostrom2024}, and \texttt{puncc} \citep{Mendil2023}, these tools are largely designed around regression and classification workflows rather than anomaly detection. The \texttt{nonconform} package was designed to address this gap:

\begin{center}
    \url{https://github.com/OliverHennhoefer/nonconform}\footnote{All descriptions in this paper are tied to the release version \texttt{v1.0.2}. Install via \texttt{pip install nonconform} or \mbox{\texttt{uv add nonconform}}, optional dependencies can be added depending on workflow specifics.}
\end{center}

\subsection{Positioning and Scope}

To the best of our knowledge, \texttt{nonconform} is currently the only Python package that treats CAD as a primary, package-level objective: anomaly detector adaptation, conformal $p$-value computation, error control procedures, covariate-shift-aware weighting, and online-capable exchangeability martingales in one coherent API.

\subsection{Core API and Modular Architecture}

The central entry point is the \texttt{ConformalDetector} meta-estimator, built in a scikit-learn-compatible style \citep{Buitinck2013}. Its core lifecycle is

\begin{center}
    \texttt{.fit()} $\rightarrow$ \texttt{.select()},
\end{center}

with optional \texttt{.compute\_p\_values()} and \texttt{.score\_samples()} for custom analysis workflows and \texttt{.calibrate()} for detached calibration of wrapped, already-fitted models.

\subsubsection{Core Modules and Responsibility Boundaries}

The architecture separates concerns across modules:
\begin{itemize}
    \item \texttt{nonconform.adapters}: detector adaptation and score-direction normalization.
    \item \texttt{nonconform.resampling}: conformalization strategies.
    \item \texttt{nonconform.scoring}: $p$-value computation.
    \item \texttt{nonconform.weighting}: estimator-based importance weighting.
    \item \texttt{nonconform.fdr}: weighted FDR selection primitives.
    \item \texttt{nonconform.martingales}: sequential exchangeability-evidence processes.
    \item \texttt{nonconform.structures}: \texttt{AnomalyDetector} protocol and result container.
\end{itemize}

\subsubsection{Conformalization Strategy}

The package provides three strategy families in \texttt{nonconform.resampling}:
\begin{center}
    \textbullet~ \texttt{Split()} \hspace{2em} 
    \textbullet~ \texttt{CrossValidation()} \hspace{2em} 
    \textbullet~ \texttt{JackknifeBootstrap()}
\end{center}
For resampling methods, \texttt{mode} (\texttt{plus} vs.\ \texttt{single\_model}) controls the model retention--validity tradeoff. \texttt{CrossValidation.jackknife()} implements leave-one-out resampling.

\subsubsection{Estimation Strategy}

On top of calibrated scores, the package offers three estimator classes:

\begin{center}
    \textbullet~ \texttt{Empirical()} \hspace{2em} 
    \textbullet~ \texttt{ConditionalEmpirical()} \hspace{2em} 
    \textcolor{gray}{\textbullet~ \texttt{Probabilistic()}}\footnote{\texttt{Probabilistic()} uses [weighted] KDE rather than empirical ranks. It is not conformal and has no finite-sample guarantee; asymptotic validity requires additional regularity assumptions.}
\end{center}

These are exposed in \texttt{nonconform.scoring} and integrated into \texttt{ConformalDetector} through the \texttt{estimation} argument. Detached calibration is only implemented for \texttt{Split()}.

\subsection{Compatibility \& Interoperability}

The polarity of processed anomaly scores is critical for compatibility over different detector implementations. Internally, \texttt{nonconform} standardizes to 

\[
\text{higher score} \rightarrow \text{more anomalous}.
\]
The \texttt{score\_polarity} interface supports explicit conventions, strict \texttt{"auto"} inference for recognized detector families, and implicit defaults when omitted.

\paragraph{PyOD.}
\texttt{nonconform} supports \texttt{PyOD} \citet{Zhao2019} detectors through adapter-based compatibility, allowing direct use of established one-class detectors (e.g., \texttt{IForest}, \texttt{LOF}).

\begin{warningbox}
Some \texttt{PyOD} detectors (e.g., \texttt{COPOD}, \texttt{ECOD}, \texttt{LOCI}, \texttt{SOS}, \texttt{COF}) are not compatible with conformal workflows, as their scores are not generated by a fixed function post-training. As a result, conformal \(p\)-values and FDR guarantees are not valid.
\end{warningbox}

\paragraph{Scikit-learn.} \texttt{nonconform} follows scikit-learn estimator conventions (\texttt{fit}, \texttt{get\_params}, \texttt{set\_params}), enabling straightforward integration into existing model-selection and pipeline code. In particular, \texttt{ConformalDetector} supports standard estimator semantics while adding conformal calibration and decision control through \texttt{select()}.

\paragraph{Custom.} Custom detector implementations can be integrated via the \texttt{AnomalyDetector} protocol. This keeps the detector layer replaceable while preserving a fixed conformal interface for calibration, $p$-value computation, and downstream FDR-controlled selection.

\subsection{Error Control Under Batch and Shift}

For standard batch usage, the recommended one-step API is \texttt{ConformalDetector.select()}, which computes conformal $p$-values and applies BH-style control in the unweighted mode. This avoids brittle manual thresholding and keeps the decision pipeline statistically explicit.

For covariate shift, \texttt{nonconform} uses estimator-based weighting via \texttt{weight\_estimator}, for example \texttt{logistic\_weight\_estimator()}. In weighted mode, \texttt{select()} dispatches to weighted conformalized selection via \texttt{weighted\_false\_discovery\_control()}.

\subsection{Sequential Monitoring}

Beyond pointwise decisions, \texttt{nonconform} supports streaming evidence monitoring through \texttt{PowerMartingale}, \texttt{SimpleMixtureMartingale}, and \texttt{SimpleJumperMartingale}. Alarm behavior is configured through \texttt{AlarmConfig} thresholds (Ville, CUSUM, Shiryaev--Roberts) and exposed in \texttt{MartingaleState.triggered\_alarms}. This design keeps sequential evidence processes separate from cross-hypothesis FDR selection logic.

\subsection{Reliability, Reproducibility, and Documentation}

The repository follows a layered validation design across unit, integration, and end-to-end tests, and CI workflows for multi-version testing coverage, strict documentation builds, and wheel/post-publish smoke checks. \href{https://oliverhennhoefer.github.io/nonconform/}{Documentation} includes API references plus executable examples for \texttt{scikit-learn}, \texttt{PyOD}, custom detectors, weighted workflows, detached calibration, conditional calibration, and martingales in greater detail:

\begin{center}
    \url{https://oliverhennhoefer.github.io/nonconform/}
\end{center}

Overall, \texttt{nonconform} contributes a robust implementation of CAD by making statistical assumptions explicit, API boundaries modular, and evaluation workflows reproducible.

\section{Conclusion}
\label{sec:conclusion}

The \texttt{nonconform} package implements a statistical decision layer for anomaly detection. By separating scoring from calibration, selection, weighting, and sequential evidence accumulation, it makes the assumptions behind each decision explicit. The resulting workflow replaces heuristic thresholds with conformal \textit{p}-values that support FDR-controlled discovery sets under the corresponding assumptions, or martingale-based evidence measures for shifts in data streams that, e.g., inform retraining. The central message is simple: reliable anomaly detection is not obtained from better scores and more complex models or workflows alone, but from decision rules whose statistical guarantees enable model-free and finite-sample valid quantification of uncertainty.

We plan to gradually expand \texttt{nonconform}'s use cases and invite open-source maintainers, CAD enthusiasts, and authors of relevant works to implement new methods, making applications and research in this field more accessible.

\acks{This work was conducted as part of a research project funded by the German Federal Ministry for Economic Affairs and Climate Action (BMWK) under grant number 01MV23020A.}

\bibliography{references}

@article{Benjamini1995,
  author  = {Benjamini, Yoav and Hochberg, Yosef},
  title   = {Controlling the False Discovery Rate: A Practical and Powerful Approach to Multiple Testing},
  journal = {Journal of the Royal Statistical Society: Series B (Methodological)},
  year    = {1995},
  volume  = {57},
  number  = {1},
  pages   = {289--300},
  doi     = {10.1111/j.2517-6161.1995.tb02031.x},
}

@book{Vovk2005,
  author    = {Vovk, Vladimir and Gammerman, Alexander and Shafer, Glenn},
  title     = {Algorithmic Learning in a Random World},
  publisher = {Springer},
  address   = {New York, NY},
  edition   = {1},
  year      = {2005},
  doi       = {10.1007/b106715},
}

@inproceedings{Hennhoefer2024,
	title        = {Leave-One-Out-,  Bootstrap- and Cross-Conformal Anomaly Detectors},
	author       = {Hennh\"{o}fer,  Oliver and Preisach,  Christine},
	year         = 2024,
	month        = dec,
	booktitle    = {2024 IEEE International Conference on Knowledge Graph (ICKG)},
	publisher    = {IEEE},
	pages        = {110--119},
	doi          = {10.1109/ickg63256.2024.00022},
}

@article{Vovk2015,
  author  = {Vovk, Vladimir},
  title   = {Cross-conformal predictors},
  journal = {Annals of Mathematics and Artificial Intelligence},
  year    = {2015},
  volume  = {74},
  number  = {1--2},
  pages   = {9--28},
  doi     = {10.1007/s10472-013-9368-4},
}

@article{Jin2025,
    author = {Jin, Ying and Candès, Emmanuel J},
    title = {Model-free selective inference under covariate shift via weighted conformal p-values},
    journal = {Biometrika},
    volume = {113},
    number = {1},
    year = {2026},
    month = {02},
    issn = {1464-3510},
    doi = {10.1093/biomet/asaf066},
}

@inproceedings{Liu2008,
  author    = {Liu, Fei Tony and Ting, Kai Ming and Zhou, Zhi-Hua},
  title     = {Isolation Forest},
  booktitle = {2008 Eighth IEEE International Conference on Data Mining},
  year      = {2008},
  pages     = {413--422},
  publisher = {IEEE},
  doi       = {10.1109/ICDM.2008.17},
}

@article{Schoelkopf2001,
	title        = {Estimating the Support of a High-Dimensional Distribution},
	author       = {Sch{\"o}lkopf, Bernhard and Platt, John C. and Shawe-Taylor, John and Smola, Alex J. and Williamson, Robert C.},
	year         = 2001,
	journal      = {Neural Computation},
	volume       = 13,
	number       = 7,
	pages        = {1443--1471},
	doi          = {10.1162/089976601750264965}
}

@article{Bates2023,
  author  = {Bates, Stephen and Cand{\`e}s, Emmanuel and Lei, Lihua and Romano, Yaniv and Sesia, Matteo},
  title   = {Testing for outliers with conformal p-values},
  journal = {The Annals of Statistics},
  year    = {2023},
  volume  = {51},
  number  = {1},
  pages   = {149--178},
  doi     = {10.1214/22-AOS2244},
}

@article{Laxhammar2015,
	title        = {Inductive conformal anomaly detection for sequential detection of anomalous sub-trajectories},
	author       = {Laxhammar, Rikard and Falkman, G{\"o}ran},
	year         = 2015,
	month        = jun,
	journal      = {Annals of Mathematics and Artificial Intelligence},
	publisher    = {Kluwer Academic Publishers},
	address      = {USA},
	volume       = 74,
	number       = {1--2},
	pages        = {67--94}
}

@article{Javanmard2018,
  title = {Online rules for control of false discovery rate and false discovery exceedance},
  volume = {46},
  ISSN = {0090-5364},
  DOI = {10.1214/17-aos1559},
  number = {2},
  journal = {The Annals of Statistics},
  publisher = {Institute of Mathematical Statistics},
  author = {Javanmard,  Adel and Montanari,  Andrea},
  year = {2018},
  month = Apr 
}

@article{Vovk2021,
	title        = {E-values: Calibration, combination and applications},
	author       = {Vovk, Vladimir and Wang, Ruodu},
	year         = 2021,
	month        = jun,
	journal      = {Ann. Stat.},
	publisher    = {Institute of Mathematical Statistics},
	volume       = 49,
    pages        = {1736--1754},
	number       = 3
}

@inproceedings{Vovk2003,
	title        = {Testing exchangeability on-line},
	author       = {Vovk, Vladimir and Nouretdinov, Ilia and Gammerman, Alex},
	year         = 2003,
	booktitle = {Proceedings of the Twentieth International Conference on Machine Learning},
	location     = {Washington, DC, USA},
	publisher    = {AAAI Press},
	series       = {ICML'03},
	pages        = {768--775},
	isbn         = 1577351894,
	numpages     = 8
}

@article{Shimodaira2000,
  author  = {Shimodaira, Hidetoshi},
  title   = {Improving predictive inference under covariate shift by weighting the log-likelihood function},
  journal = {Journal of Statistical Planning and Inference},
  year    = {2000},
  volume  = {90},
  number  = {2},
  pages   = {227--244},
  doi     = {10.1016/S0378-3758(00)00115-4},
}

@inproceedings{Cordier2023,
	title        = {{Flexible and Systematic Uncertainty Estimation with Conformal Prediction via the MAPIE library}},
	author       = {Cordier, Thibault and Blot, Vincent and Lacombe, Louis and Morzadec, Thomas and Capitaine, Arnaud and Brunel, Nicolas},
	year         = 2023,
	booktitle    = {Conformal and Probabilistic Prediction with Applications},
    volume       = 204,
	pages        = {549--581}
}

@inproceedings{Mendil2023,
	title = {{PUNCC}: a Python Library for Predictive Uncertainty Calibration and Conformalization},
	author       = {Mendil, Mouhcine and Mossina, Luca and Vigouroux, David},
	year         = 2023,
	booktitle    = {Conformal and Probabilistic Prediction with Applications},
	pages        = {582--601},
    volume       = 204,
	publisher = {PMLR}
}

@inproceedings{Bostrom2024,
  author    = {Bostr\"{o}m, Henrik},
  title     = {Conformal Prediction in Python with crepes},
  booktitle = {Proceedings of the Thirteenth Symposium on Conformal and Probabilistic Prediction with Applications},
  year      = {2024},
  editor    = {Vantini, Simone and Fontana, Matteo and Solari, Aldo and Bostr\"{o}m, Henrik and Carlsson, Lars},
  volume    = {230},
  series    = {Proceedings of Machine Learning Research},
  pages     = {236--249},
  publisher = {PMLR},
}

@article{Zhao2019,
  author  = {Zhao, Yue and Nasrullah, Zain and Li, Zheng},
  title   = {{PyOD}: A Python Toolbox for Scalable Outlier Detection},
  journal = {Journal of Machine Learning Research},
  year    = {2019},
  volume  = {20},
  number  = {96},
  pages   = {1--7},
}

@misc{Hennhofer2026,
	title        = {Between resolution collapse and variance inflation: Weighted conformal anomaly detection in low-data regimes},
	author       = {Hennh{\"o}fer, Oliver and Preisach, Christine},
	year         = 2026,
	month        = mar,
	copyright    = {http://arxiv.org/licenses/nonexclusive-distrib/1.0/},
	eprint        = {2603.23205},
    archiveprefix = {arXiv},
    primaryclass  = {stat.ML},
}

@article{Shaffer1995,
  author  = {Shaffer, Juliet Popper},
  title   = {Multiple hypothesis testing},
  journal = {Annual Review of Psychology},
  year    = {1995},
  volume  = {46},
  number  = {1},
  pages   = {561--584},
  doi     = {10.1146/annurev.ps.46.020195.003021},
}

@inproceedings{Fedorova2012,
  author    = {Fedorova, Valentina and Gammerman, Alex and Nouretdinov, Ilia and Vovk, Vladimir},
  title     = {Plug-in martingales for testing exchangeability on-line},
  booktitle = {Proceedings of the 29th International Conference on Machine Learning (ICML-12)},
  year      = {2012},
  editor    = {Langford, John and Pineau, Joelle},
  pages     = {1639--1646},
  publisher = {Omnipress},
}

@inproceedings{Vovk2021b,
  author    = {Vovk, Vladimir and Petej, Ivan and Nouretdinov, Ilia and Ahlberg, Ernst and Carlsson, Lars and Gammerman, Alex},
  title     = {Retrain or not retrain: conformal test martingales for change-point detection},
  booktitle = {Proceedings of the Tenth Symposium on Conformal and Probabilistic Prediction and Applications},
  year      = {2021},
  volume    = {152},
  series    = {Proceedings of Machine Learning Research},
  pages     = {191--210},
  publisher = {PMLR},
}

@inproceedings{Buitinck2013,
  author    = {Lars Buitinck and Gilles Louppe and Mathieu Blondel and
               Fabian Pedregosa and Andreas Mueller and Olivier Grisel and
               Vlad Niculae and Peter Prettenhofer and Alexandre Gramfort
               and Jaques Grobler and Robert Layton and Jake VanderPlas and
               Arnaud Joly and Brian Holt and Ga{\"{e}}l Varoquaux},
  title     = {{API} design for machine learning software: experiences from the scikit-learn
               project},
  booktitle = {ECML PKDD Workshop: Languages for Data Mining and Machine Learning},
  year      = {2013},
  pages = {108--122},
}

@misc{Shuttle,
  author       = {Feng, C. and Sutherland, A. and King, S. and Muggleton, S. and Henery, R.},
  title        = {{Statlog Project}},
  year         = {1993},
  howpublished = {UCI Machine Learning Repository},
  note         = {{DOI}: https://doi.org/10.24432/C5XS3B}
}

\end{document}